\documentclass[sigconf,final]{acmart}

\setcitestyle{square}

\usepackage[acronym]{glossaries}
\newacronym{ik}{IK}{Inverse Kinematics}

\renewcommand{\vec}[1]{\mathbf{#1}}

\usepackage{stfloats} 

\renewcommand{\vec}[1]{\mathbf{#1}}

\DeclareMathOperator{\att}{\textsc{ATT}}
\DeclareMathOperator{\mha}{\textsc{MHA}}
\DeclareMathOperator{\ffn}{\textsc{FFN}}
\DeclareMathOperator{\relu}{\textsc{ReLu}}
\DeclareMathOperator{\layernorm}{\textsc{LayerNorm}}
\DeclareMathOperator{\softmax}{\textsc{softmax}}
\DeclareMathOperator{\concat}{\textsc{concat}}

\begin{document}


\title{3D Human Pose and Shape Estimation via HybrIK-Transformer}

\author{Boris N. Oreshkin}
\email{boris.oreshkin@gmail.com}
% \affiliation{%
%   \institution{Unity Technologies}
%   \country{Canada}
% }

% \renewcommand{\shortauthors}{Voleti, Oreshkin, Bocquelet, et al.}

\begin{abstract}
HybrIK relies on a combination of analytical inverse kinematics and deep learning to produce more accurate 3D pose estimation from 2D monocular images~\cite{li2021hybrik}. HybrIK has three major components: (1) pretrained convolution backbone, (2) deconvolution to lift 3D pose from 2D convolution features, (3) analytical inverse kinematics pass correcting deep learning prediction using learned distribution of plausible twist and swing angles. In this paper we propose an enhancement of the 2D to 3D lifting module, replacing deconvolution with Transformer, resulting in accuracy and computational efficiency improvement relative to the original HybrIK method. We demonstrate our results on commonly used H36M, PW3D, COCO and HP3D datasets. Our code is publicly available \url{https://github.com/boreshkinai/hybrik-transformer}.
\end{abstract}

\maketitle

\begin{table*}[!b]
    \begin{center}
    \resizebox{\textwidth}{!}
    {%
        \begin{tabular}{l|ccc|cc|ccc}
        \toprule
        & \multicolumn{3}{c}{3DPW} & \multicolumn{2}{c}{Human3.6M} & \multicolumn{3}{c}{MPI-INF-3DHP} \\
        \cmidrule(lr){2-4} \cmidrule(lr){5-6} \cmidrule(lr){7-9}
        Method & PA-MPJPE $\downarrow$ & MPJPE $\downarrow$ & PVE $\downarrow$ & PA-MPJPE $\downarrow$ & MPJPE $\downarrow$ & PCK $\uparrow$ & AUC $\uparrow$ & MPJPE $\downarrow$ \\
        \midrule
        HMR~\cite{hmr} & 81.3 &  130.0 &  - &  56.8 &  88.0 &  72.9 &  36.5 &  124.2 \\
        CMR~\cite{cmr} & 70.2 & - & - & 50.1 & - & - & - & - \\
        \citet{pavlakos2018learning} &  - &  - &  - &  75.9 &  - &  - &  - &  - \\
        \citet{arnab2019exploiting} & 72.2 & - & - & 54.3 & 77.8 & - & - & - \\
         SPIN~\cite{spin} &  59.2 &  96.9 &  116.4 &  41.1 &  - &  76.4 &  37.1 &  105.2 \\
        I2L~\cite{i2l}$^*$ & 58.6 & 93.2 & - & 41.7 & 55.7 & - & - & - \\
         Mesh Graphormer~\cite{lin2021mesh} \textit{w. 3DPW} &  45.6 &  74.7 &  87.7 &  41.2 &  34.5 &  - &  - &  - \\
        PARE~\cite{kocabas2021pare} \textit{w. 3DPW} & 46.4 & 74.7 & 87.7 & - & - & - & - & - \\
        \midrule
        HybrIK~\cite{li2021hybrik} & {48.8} & {80.0} & {94.5} & 34.5 & 54.4 & {86.2} & {42.2} & {91.0} \\
        Ours (HybrIK-Transformer ResNet-34) & 47.6 & 75.7 & 91.4 & 34.9 & 52.5 & 88.3 & 48.4 & 88.3 \\
        Ours (HybrIK-Transformer HrNet-48) & \textbf{43.4} & \textbf{73.6} & \textbf{87.3} & \textbf{29.8} & \textbf{48.8} & \textbf{89.3} & \textbf{49.0} & \textbf{85.2} \\
        \midrule
        HybrIK~\cite{li2021hybrik} \textit{w. 3DPW} & 45.0 & 74.1 & 86.5 & 33.6 & {55.4} & 87.5 & {46.9} & {93.9} \\
        Ours (HybrIK-Transformer ResNet-34) \textit{w. 3DPW} & {46.0} & {74.9} & {88.1} & {34.6} & 50.2 & {86.7} & 47.6 & 90.2 \\
        Ours (HybrIK-Transformer HrNet-48) \textit{w. 3DPW}  & \textbf{42.3} & \textbf{71.6} & \textbf{83.6} & \textbf{29.5} & \textbf{47.5} & \textbf{88.6} & \textbf{48.9} & \textbf{86.2} \\
        \bottomrule
        \end{tabular}
        
        % ##### gt 3dpw err: {'MPJPE': 75.73766629269811, 'PA-MPJPE': 47.59498667785608, 'PVE': 91.37223912871998} #####
        % ##### gt 3dhp err: {'MPJPE': 88.34067231194521, 'PA-MPJPE': 64.25475505989111, 'PCK': 88.25093167701864, 'AUC': 48.36802244039271} #####
        % ##### gt h36m err: {'PA-MPJPE': 34.87701486731242, 'MPJPE': 52.5489746919412} #####
        
    }
    \end{center}
    % \vspace{-3 mm}
    \caption{Benchmark of state-of-the-art models on 3DPW, Human3.6M and MPI-INF-3DHP datasets. ``$*$'' denotes the method is trained on different datasets. ``-'' shows the results that are not available.}
    \label{tab:key_results}
    \vspace{-3mm}
\end{table*}

% \section{Introduction}

\section{Method}

We use the same backbone and analytical IK approach (components 1 and 3 outlined in abstract) as original HybrIK~\cite{li2021hybrik}. However, we note that the bulk of the processing happens in component 2, the 3D lifting/keypoint estimation block, inherited from~\cite{integral}. This consists of three deconvolution layers followed by a $1 \times 1$ convolution to generate the 3D joint heatmaps. The soft-argmax operation is used to obtain 3D pose from the heatmap in a differentiable manner. 

This operation is very costly both in terms of parameter counts and GPU memory consumption. We propose to replace it with Transformer blocks conveniently described in terms of the original primitives~\cite{vaswani2017attention}. The attention unit is defined as:
\begin{align} \label{eqn:attention}
    \att(\vec{Q}, \vec{K}, \vec{\vec{V}}) = \softmax(\vec{Q} \vec{K}^T / \sqrt{d}) \vec{V},
\end{align}
where $d$ is model-wide hidden dimension width. The multi-head attention unit is defined as:
\begin{align}
    \vec{h}_i &= \att(\vec{Q} \vec{W}^{Q}_{i}, \vec{K} \vec{W}^{K}_{i}, \vec{\vec{V}} \vec{W}^{V}_{i}), \nonumber \\
    \mha(\vec{Q}, \vec{K}, \vec{\vec{V}}) &= \concat(\vec{h}_1, \ldots, \vec{h}_h) \vec{W}^{O}. \nonumber
\end{align}
Transformer also uses feed-forward network $\ffn$, which is a simple multi-layer perceptron.

Our proposed 3D keypoint estimation procedure consists of $L$ Transformer blocks and each block $\ell \in [1, L]$ performs the following computation. First, encode 2d image features:
\begin{align}
    \vec{e}_{\ell}^{\textrm{2d}}  &= \mha_{\ell}^{\textrm{2d}}(\vec{e}_{\ell-1}^{\textrm{2d}}, \vec{e}_{\ell-1}^{\textrm{2d}}, \vec{e}_{\ell-1}^{\textrm{2d}}), \nonumber \\
    \vec{e}_{\ell}^{\textrm{2d}} &= \relu(\layernorm_{\ell}^{\textrm{2d}}(\vec{e}_{\ell}^{\textrm{2d}} + \vec{e}_{\ell-1}^{\textrm{2d}})), \nonumber \\
    \vec{e}_{\ell}^{\textrm{2d}} &= \ffn_{\ell}^{\textrm{2d}}(\vec{e}_{\ell}^{\textrm{2d}}). \nonumber 
\end{align}
Here $\vec{e}_{0}^{\textrm{2d}}$ is the source input of transformer, which is a tensor of flattened 2D image features derived from the output of ResNet backbone, summed with learnable 2D position encoding and projected to internal transformer width $d$. 

Second, encode target 3D templates:
\begin{align}
    \vec{e}_{\ell}^{\textrm{3d-t}}  &= \mha_{\ell}^{\textrm{3d}}(\vec{e}_{\ell-1}^{\textrm{3d}}, \vec{e}_{\ell-1}^{\textrm{3d}}, \vec{e}_{\ell-1}^{\textrm{3d}}), \nonumber \\
    \vec{e}_{\ell}^{\textrm{3d-t}} &= \relu(\layernorm_{\ell}^{\textrm{3d}}(\vec{e}_{\ell}^{\textrm{3d-t}} + \vec{e}_{\ell-1}^{\textrm{3d}})), \nonumber 
\end{align}
Here $\vec{e}_{0}^{\textrm{3d}}$ is the target input of transformer, which is a $28 \times d$ tensor containing templates for the following desired outputs: (i) 24 3D keypoints, (ii) 23 twist angles $\Phi$, (iii) SMPL shape parameters $\beta$. Each template consists of the sum of the corresponding joint embedding and the output type embedding. There are 24 joint embeddings for each SMPL joint and 3 output type embeddings: 3D keypoint, twist angle, SMPL shape.

Finally, decode 3D outputs from encoded 2D inputs and encoded 3D templates:
\begin{align}
    \vec{e}_{\ell}^{\textrm{3d}}  &= \mha_{\ell}^{\textrm{2d-3d}}(\vec{e}_{\ell}^{\textrm{3d-t}}, \vec{e}_{\ell}^{\textrm{2d}}, \vec{e}_{\ell}^{\textrm{2d}}), \nonumber \\
    \vec{e}_{\ell}^{\textrm{3d}} &= \relu(\layernorm_{\ell}^{\textrm{2d-3d}}(\vec{e}_{\ell}^{\textrm{3d}} + \vec{e}_{\ell}^{\textrm{3d-t}})), \nonumber \\
    \vec{e}_{\ell}^{\textrm{3d}} &= \ffn_{\ell}^{\textrm{3d}}(\vec{e}_{\ell}^{\textrm{3d}}). \nonumber 
\end{align}
Transformer output $\vec{e}_{L}^{\textrm{3d}}$ is a $28 \times d$ tensor. Its leading 24 elements are linearly projected into 3D joint keypoints, next 23 positions are linearly projected into cosine and sine of the twist angle and the final element is linearly projected into 10-dimensional SMPL shape $\beta$.

A small note on training methodology is in order. At train time we employ augmentation of $\vec{e}_{0}^{\textrm{2d}}$ by sampling a subset of ResNet patches to pass to the transformer input. We implement it using multinomial sampling without replacement guided by a randomly generated integer specifying the number of patches to retain in each batch. At inference time, all ResNet patches are passed to transformer input. We observed a small but consistent positive regularizing effect resulting from this augmentation.

\section{Results}

The following datasets are used in our empirical studies. We use the same protocols and dataset preprocessing as~\citet{li2021hybrik}. \textbf{MSCOCO}~\cite{cocodataset} is a large-scale in-the-wild 2D human pose dataset used only for training. \textbf{3DPW}~\cite{vonMarcard2018} is a challenging outdoor benchmark for 3D pose and shape estimation used both for training and evaluation. \textbf{MPI-INF-3DHP}~\cite{mehta2017monocular} consists of both constrained indoor and complex outdoor multi-camera 3D scenes. We use its train set for training and evaluate on its test set. \textbf{Human3.6M}~\cite{IonescuSminchisescu11} is an indoor 3D pose estimation benchmark used for training and evaluation. We use 5 subjects (S1, S5, S6, S7, S8) for training and 2 subjects (S9, S11) for evaluation. 

\textbf{Key accuracy results} are presented in Table~\ref{tab:key_results}. Specifically, comparing our rest against original HybrIK we see consistent improvement when trained on \textbf{MSCOCO}, \textbf{Human3.6M}, \textbf{MPI-INF-3DHP}. The results of our algorithm when trained on these datasets plus \textbf{3DPW} are more mixed (which is also the case for original HybrIK --- some of its metrics decline when trained with \textbf{3DPW}), but we see that the inclusion of \textbf{3DPW} improves our results on \textbf{3DPW} itself, which is intuitively plausible.

\textbf{Computational efficiency.} Based on the ResNet-34 backbone, proposed architecture fits in 28GB GPU memory total on 4 NVIDIA M40 GPUs and batch size 256 (64 per GPU). The training speed is 63 min per epoch. The original HybrIK architecture in the same setup fits in 60GB GPU memory total and trains at 82 min per epoch. This is two times less GPU memory footprint and 23\% training speed improvement. Moreover, HybrIK-Transformer fits on 2xP100 NVIDIA GPUs occupying total 24GB of GPU memory with 2x128 batch size, whereas the original HybrIK does not fit this hardware setup. So, the proposed modification can be trained in some very resource limited setups, unlike the original HybrIK architecture.

\textbf{Hyperparamter settings} used to report results in Table~\ref{tab:key_results} are as follows. The architecture is implemented in PyTorch. We train for 200 epochs using Adam optimizer and inverse square root learning rate schedule with max learning rate 0.0005 and warmup period of 4000 batches. The models of last 10 epochs are averaged and the results of this average model are reported in Table~\ref{tab:key_results}. Following~\cite{li2021hybrik} we use ResNet-34~\cite{resnet} backbone, initialized with ImageNet pre-trained weights, the analytical IK pass converting 3D keypoint predictions and twist angles into full pose and root joint prediction from RootNet~\cite{moon}. The Transformer part uses $L=6$ blocks, $h=8$ heads, model width $d=512$, dropout 0.1, FFN has 3 layers, all of width $d=512$. We also trained and evaluated the variant of the architecture based on the HrNet-48 backbone~\cite{sun2019deep}. To integrate the HrNet backbone with transformer, we discard HrNet's mesh generation step as well as average pooling. Thus, we take only $8\times8\times2048$ raw features obtained from HrNet's final feature layer and feed them directly into transformer, exactly the way we do with the output of the ResNet-34 backbone. HrNet variant is trained with batch size $4\times 56$ to fit into 4 GPUs.

\bibliographystyle{ACM-Reference-Format}
\bibliography{ref}

% \clearpage

\end{document}